\documentclass[11pt]{article}
\usepackage[margin=1in]{geometry}
\usepackage[T1]{fontenc}
\usepackage{lmodern}
\usepackage{amsmath}
\usepackage{graphicx}
\usepackage{booktabs}
\usepackage{tabularx}
\usepackage{array}
\usepackage{xcolor}
\usepackage[numbers,sort&compress]{natbib}
\usepackage{url}
\usepackage[hidelinks]{hyperref}

\newcommand{\taub}{$\tau$-bench}
\newcommand{\tautwo}{$\tau^2$-bench}

\title{Silent Instrument Errors Reorder Agent Leaderboards}
\author{Yangze Liu\\Shandong University}
\date{}

\begin{document}
\maketitle

\begin{abstract}
On a public agent leaderboard, a general-purpose agent trails a scaffold built for an airline customer-service benchmark by 13 points on the tasks that touch its database, and a reader could conclude that generality costs accuracy. The gap comes from the grader. The general agent often ends its conversations in a way that never triggers grading, so only about half of its episodes were scored and the rest were stored as failures. Grading the ungraded half under the benchmark's own rule closes the pooled gap, without rerunning any agent. We audit two carefully maintained public leaderboards, HAL and \tautwo{}, from their published artifacts alone and find three more instrument errors. A cost column counts nested model calls twice, unevenly across rows. Archived trajectories record grader-written actions as agent actions. Re-grading saved trajectories under a newer evaluator is itself a new measurement: a correct and announced fix reorders a leaderboard, and a mode check that misreads four files collapses their scores by 59 to 82 points while printing no error. Each produced a plausible number, and the HAL maintainers confirmed their three within five weeks. Together the cases show that the infrastructure of agent evaluation, meaning the code that grades, logs, replays and versions an evaluation, is part of the measurement and fails silently, so scores and orderings can move with no visible sign. These failure modes follow from how agent evaluation pipelines are built, and those pipelines change as fast as the agents they measure, so we expect more of them elsewhere. We propose cheap checks that expose the instrument: report the graded fraction, aggregate cost from root spans, tag the provenance of every recorded action, and pin evaluator versions with a published re-grade diff.
\end{abstract}

\section{Introduction}

The Holistic Agent Leaderboard (HAL) evaluates agents on TAU-bench Airline, a customer-service benchmark in which an agent changes flight reservations for a simulated customer \citep{kapoor2025hal,yao2024taubench}. HAL has run the same models under Tool-Calling, a scaffold written for the benchmark, and under its own Generalist agent. In the results HAL publishes for the nine models run under both, on the 41 of 50 tasks whose reference solution queries or changes the database, the Generalist trails by 13 points and comes out behind for eight of them. A reader would conclude that a general-purpose agent pays a large price on a task-specific benchmark. That conclusion was produced by the grader. TAU-bench computes the reward only when an episode formally ends, and the Generalist often closes its episodes in a way that does not trigger that ending, so only about half of its episodes were ever graded and the rest were recorded as failures. When the ungraded episodes are graded offline under the same rule, the pooled gap between the two scaffolds closes to a tie and most per-model gaps shrink to a few points, while Tool-Calling, whose episodes had already been graded, barely moves (Figure~\ref{fig:regrade}).

This gap comes from the leaderboard built by the people who set the agenda for rigorous agent evaluation. Kapoor et al.\ argued that agent benchmarks should report cost alongside accuracy and that evaluation practice lacked the standardization needed for reproducible results \citep{kapoor2024agents}. HAL is the infrastructure that answered that call, with a shared harness meant to remove common implementation bugs and full trajectories released for anyone to inspect \citep{kapoor2025hal}. If the instrument could fail silently here, carelessness is the wrong explanation. The error lives in the way agent evaluation pipelines are put together, and other pipelines are built the same way.

The pipelines also change constantly. Agent scaffolds, tracing libraries, simulators and evaluators all ship new versions on the timescale of weeks, and a public leaderboard accumulates rows produced under many of those versions. Each version is a slightly different instrument. A model submitted in February and a model submitted in July may have been scored by different grading code, and the leaderboard presents both numbers in one column as if a single ruler had produced them. The field has learned to ask which model version and which scaffold produced a score. It rarely asks which instrument did.

Instrument errors are hard to see because a score carries no trace of how it was made. A grader that skips an episode returns a zero, which looks like a failure. A cost aggregator that counts a call twice returns a dollar figure in the range of other dollar figures. A re-grade that silently drops part of an environment returns a pass rate between zero and one. Benchmark audits examine whether tasks and reward rules are sound \citep{zhu2025abc}, and a recent survey from the HAL group counts harnesses and autograders among the surfaces that agent evaluation must now scrutinize \citep{chen2026emerging}. Yet the run-time code that applies those rules to a particular trajectory, logs it, prices it and stores it is rarely re-derived after the leaderboard is published.

This paper re-derives the run-time instrument of two public leaderboards, HAL and \tautwo{}, from their released artifacts. What we find supports one conclusion: the instrument must be audited as part of the measurement, and leaderboards should publish the few numbers that let a reader audit it. Section~\ref{sec:exposed} explains why agent evaluations are especially exposed. Section~\ref{sec:cases} presents the four errors we found, each from public artifacts without running an agent. Even a correct and announced evaluator fix reorders a leaderboard: a documented \tautwo{} re-grade moves one model from ninth place to a tie for fifth. Section~\ref{sec:pattern} reads the four cases together. Section~\ref{sec:todo} proposes checks cheap enough to adopt now, and Section~\ref{sec:alt} discusses what the cases mean for fixes, coarse comparisons, disclosure and open releases. The HAL maintainers reproduced the three HAL errors on their own archive and added disclosures to the leaderboard within five weeks of our report \citep{hf2026discussion9}. That response is disclosure at its best, and the archived numbers are still as they were.

\section{Why agent evaluation is exposed}
\label{sec:exposed}

The four cases in Section~\ref{sec:cases} support one claim. \emph{The code that decides when an agent is graded, what is logged and priced, how a trajectory is replayed and which evaluator version applies is part of the measurement. It changes as fast as the agents, its errors produce plausible numbers rather than failures, and so leaderboards should publish checks on the instrument alongside the scores.}

Static benchmarks are less exposed to this. A question-answering benchmark compares a string to a reference once, in one function, and the grading code rarely depends on what the model did before answering. Agent evaluation differs in four places, and each of them gives an error room to hide.

Grading is triggered by agent behaviour. In an interactive benchmark the grader runs when the episode ends, and the episode ends when the agent or a simulated user does something specific, such as calling a terminating tool or emitting a stop token. A scaffold that ends conversations differently from the one the benchmark authors had in mind can therefore change whether grading happens at all, and a missing grade is usually stored as a failure.

Logging has layers. A single model call passes through a scaffold, a wrapper library and a provider client, and modern tracing instruments each layer. Whether a cost or token total is right depends on which layers the aggregator sums, a choice that is invisible in the total and that varies with the client path each run happens to take.

Replay writes into records. To decide whether an agent reached the right end state, a stateful grader often replays the gold actions through the same environment object the agent used. Anything the environment records during that replay sits in the same lists and logs as the agent's own actions, and the trajectory archive inherits it.

Evaluators and archives move at different speeds. Benchmarks release new grading code and new task criteria while old results stay on the leaderboard and old trajectories stay in storage. Re-grading a saved trajectory under new code is offered as a way to keep results comparable, but it is a new measurement whose correctness depends on the new code understanding the old file. Once a harness is archived, as HAL's repository was in July 2026, corrections tend to arrive as notes beside numbers that stay as they were.

\section{Four instrument errors}
\label{sec:cases}

We re-derived parts of two public leaderboards from what they publish: HAL's encrypted trace archives on Hugging Face, which its own published script decrypts, and \tautwo{}'s leaderboard and repository trajectories \citep{kapoor2025hal,barres2025tau2}. No agent was run and no model was called. The HAL numbers were recomputed a second time with independently written code, and every number reproduces from the released scripts \citep{liu2026release}. Table~\ref{tab:cases} sets the four cases side by side, and Appendix~\ref{app:method} gives the procedures.

\begin{table}[t]
\centering
\small
\begin{tabularx}{\textwidth}{>{\raggedright\arraybackslash}p{2.1cm}>{\raggedright\arraybackslash}X>{\raggedright\arraybackslash}X>{\raggedright\arraybackslash}p{3.1cm}}
\toprule
Case & What the score claimed & What the instrument did & How much it moved \\
\midrule
HAL cost & Dollars spent by the agent in a run & Summed usage over a wrapper span and its child span, which carry the same response & Exactly $2\times$ in 70 of 134 runs, $1\times$ in 46 \\
HAL Generalist grading & Accuracy of the Generalist scaffold & Graded only episodes the agent transferred or the simulated user closed, stored the rest as failures & 0.30 to 0.43, closing a 13-point gap to Tool-Calling \\
HAL trajectories & Actions taken by the agent & Wrote the grader's gold replay into the agent's action list & Present in 68\% of 2,334 airline records \\
\tautwo{} re-grade & The same trajectory scored again & Applied new criteria or rebuilt the environment without the user's tools & $-7$ to $+2.5$ points on airline, up to $-82$ on telecom \\
\bottomrule
\end{tabularx}
\caption{The four cases. In none of them did the pipeline raise an error, and each produced a number in the range a reader would expect.}
\label{tab:cases}
\end{table}

\subsection{A cost column that counts some calls twice}

HAL's cost column is the realization of the cost-aware evaluation Kapoor et al.\ called for, and for many rows it reports double the money that was spent. The harness logs model calls with a tracing library in which a wrapper call and the provider-client call it makes appear as a parent span and a child span with the same response and the same usage block. The aggregation sums usage over every span, so wherever the nesting occurs the tokens and dollars are counted twice. In just over half of the runs with logs the reported tokens are exactly twice the root-span tokens, and 44 of the 94 leaderboard rows we could match to a logged archive carry a doubled cost in the archive's own results field. The doubling follows the client path a run takes, which depends on the scaffold and on the model's provider. On SciCode all ten zero-shot runs are clean while most runs of the other two scaffolds sit at about twice their cost, so one benchmark's cost axis mixes doubled and undoubled rows.

A uniform factor of two would only rescale the axis. An uneven one moves points relative to each other, and the cost-accuracy frontier that HAL exists to draw changes shape. One row shows how invisible this is: the Tool-Calling scaffold with Claude Opus 4.1 at high reasoning is listed at \$149.98 for a run whose root-span cost is \$74.99. Both figures are believable prices for fifty airline tasks, and nothing in the row tells a reader which one is real. The HAL paper removes ``any redundant calls related to weave logging'' when it prepares transcripts for its own log analysis \citep{kapoor2025hal}, so the duplication was handled in one part of the pipeline and not in the part that computed the cost column, the kind of gap that opens when components are written by different people at different times.

\subsection{An agent that is graded half the time}

The Generalist's poor TAU-bench ranking is largely a measure of how its episodes end. In TAU-bench the reward is computed only when an episode is done, which happens when the agent calls a terminating tool or when the simulated user answers a message with a stop token \citep{yao2024taubench}. The Generalist runs its own loop to completion, sends one closing message to the user and exits. If it handed the customer to a human agent along the way, that transfer is a terminating tool and the grader runs. Otherwise the closing message is unsolicited, the simulator usually does not answer it with the stop token, and no further step follows, so the database is never compared and the episode is stored with reward zero. Separating the grader's actions from the agent's own tool calls in the raw logs (Section~\ref{sec:gold}) shows which episodes were graded. Tool-Calling was graded in 98\% of its episodes and the Generalist in 51\%. Its ungraded episodes contain no ending event at all and end on a message to the user, at a median of nine actions, far below any step limit.

Grading those episodes offline closes the Generalist's pooled deficit against Tool-Calling. With the agent's own actions replayed through the pinned environment and the benchmark's own reward function, the Generalist's pooled accuracy on those 41 tasks rises from 0.30 to 0.43, and Tool-Calling barely moves because its episodes had already been graded. Over the nine models run under both scaffolds, a published gap of 13 points in Tool-Calling's favour becomes a tie. On this leaderboard the difference between the two scaffolds was a difference in whether the grader ran.

Part of the correction cannot come from doing nothing. Most of the pooled gain comes from tasks whose gold only reads the database, which the benchmark's rule passes whenever the episode leaves the database as it was, exactly as it passes the graded episodes of the other scaffold. On tasks whose gold requires a database write an untouched database fails, and the Generalist still rises from 0.28 to 0.33. Each of the 23 ungraded episodes that pass on those tasks contains the required write among the agent's own tool calls. These are solved tasks recorded as failures.

\begin{figure}[t]
\centering
\includegraphics[width=0.82\textwidth]{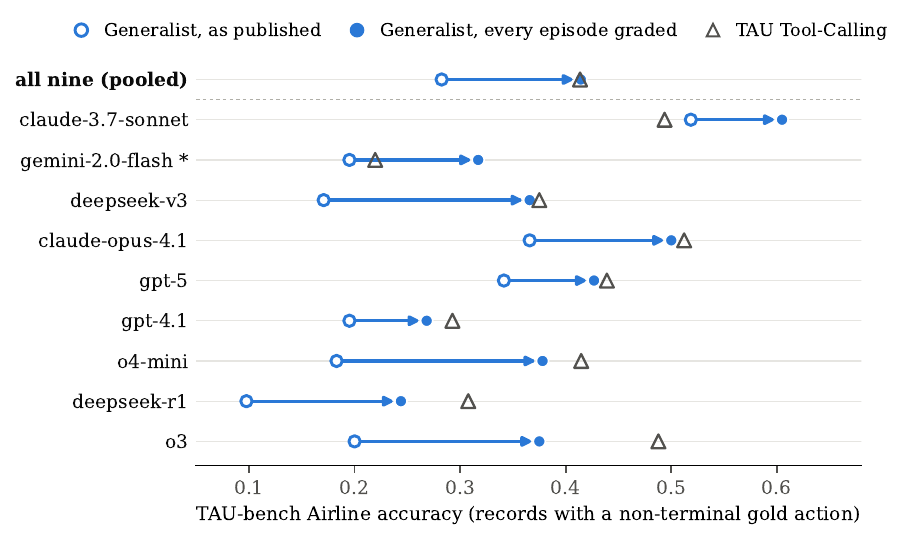}
\caption{Generalist against Tool-Calling on HAL TAU-bench Airline, for the nine models run under both, on records whose gold contains at least one non-terminal action, pooled over each model's runs. Arrows run from the Generalist's published accuracy to its accuracy when every episode is graded. Triangles are Tool-Calling after the same re-grading, which moves it by at most 2.5 points. The top row pools all nine models, where the published gap closes to a tie. The asterisk marks the one model on which the two scaffolds change order. Each per-model row rests on 40 to 82 Generalist records. Few-Shot, which HAL withdrew, is in Table~\ref{tab:permodel}.}
\label{fig:regrade}
\end{figure}

Per model, the correction moves every comparison toward the Generalist and changes the order of the two scaffolds for one model, gemini-2.0-flash (Figure~\ref{fig:regrade}). The Generalist trails Tool-Calling for eight of the nine models as published and for seven once every episode is graded. Run by run, grading every episode reverses 48 of the 351 pairwise orderings among the 27 archived Generalist and Tool-Calling runs on these 41 tasks. A model's cell holds one or two runs of about 41 records each, so the per-model gaps that end within a few points of zero are not resolved by these data, and the claim we rest on them is the direction of the shift. HAL also ran a third scaffold, Few-Shot, and withdrew its results after its own log analysis found benchmark tasks among the demonstrations \citep{kapoor2025hal}. The comparison with its archived traces moves the same way, as Appendix~\ref{app:method} reports. The scaffolds still differ in design, since they run under different step budgets and Tool-Calling places the airline policy in the system prompt while the Generalist must open a file to read it, so the corrected comparison is between scaffolds as built, now graded alike. What it no longer supports is the reading that a general agent pays a large price on this benchmark.

\subsection{Trajectories that contain the grader's actions}
\label{sec:gold}

The archived trajectories record actions the agent never took. TAU-bench's environment appends every executed action to a list, and to compute the reward the grader resets the data and replays the gold actions through the same step function, which appends them to the same list. HAL stores that list as the agent's taken actions. Aligning each record with the agent's own tool calls in the raw logs finds the grader's actions in 68\% of the 2,334 public airline records. They usually sit at the end. In Generalist episodes that ended with a transfer to a human agent they sit in the middle, because the harness appends its closing message after the grader has run. In 58 records the agent made no tool call at all while the record displays the complete gold tool sequence.

The injected actions change what the records say about the agent. Replaying the stored actions verbatim reproduces the recorded reward in 70\% of the records whose gold has at least two non-terminal actions, and removing the grader's actions first raises agreement to 92\%. Anyone who counts actions or fine-tunes on these traces is reading partly grader behaviour, and the contamination is strongest exactly where the agent did least. This case matters beyond HAL because trajectory archives are now treated as a research product in their own right \citep{kapoor2025hal,kirgis2026log}. A reader who strips only the end of each record misses the copies in the middle, and nothing in the record warns that they are there. A log is evidence about the agent only if the pipeline records who wrote each line.

\subsection{Re-grading that is not version stable}

\tautwo{} offers re-grading as the way to keep old trajectories comparable under new grading code, and that operation is itself a new measurement with its own errors \citep{barres2025tau2,tau2repo}. We re-graded public trajectory files under four evaluator builds and crossed evaluator code with task data. The results fall into three kinds, which differ in how visible they were.

The banking re-grade in release 1.0.1 was announced, and it still reordered the leaderboard. The changelog warned that banking scores were not comparable across the release and said that its grading fixes would move scores only upward, with one corrected gold value as the exception that could fail a task. Our re-grade of the paired simulations finds no pass that became a fail, and all of the movement comes from evaluator code with none from task data. Across the official leaderboard the re-grade still reverses 4 of the 136 pairwise orderings among its 17 submissions and moves gemini-3-flash from ninth place to a tie for fifth. This is the case done well, and it shows that even a documented evaluator change is a ranking event.

The airline re-grade was listed and not flagged. Release 1.0.0 changed the evaluation criteria of 14 of the 50 airline tasks in substance while leaving the database and the policy unchanged, and its changelog lists the task fixes without a warning about comparability. Applied through the documented re-grading command to four older airline files, the new criteria flip 77 of 800 simulations in both directions, move per-model scores from $-7$ to $+2.5$ points and reverse the order of gpt-4.1-mini and claude-3-7-sonnet. A reader who compares a re-graded file with one that was not re-graded is comparing two instruments.

The telecom re-grade raised no error. The re-grading command decides whether a file was recorded in solo mode by matching one agent name, and four repository files recorded with a closely related solo agent fall through to the non-solo path. The replay environment is then built without the user-side tools, every replayed user call returns a not-found message, and under the lenient replay introduced in 1.0.1 the command warns per call and continues against a state that never received those changes. The four scores fall by 59 to 82 points, and the command prints a number. A companion file recorded with the recognised solo agent re-grades to its recorded score, which isolates the cause to the mode check. A community pull request fixes the check \citep{tau2bench2026pr508}, but any score produced in the meantime looks like any other.

\subsection{What happened after disclosure}

We reported the three HAL items to the maintainers and on the trace dataset on 3 September 2026, filed the \taub{} root cause as an issue, and filed the \tautwo{} items as an issue \citep{hf2026discussion9,taubench2026issue92,tau2bench2026issue502}. On 7 October a HAL maintainer replied that the team had reproduced all three on its own archive and added disclosures to the leaderboard, which are dated 5 October. As of 3 November 2026 the cost note on each benchmark page says that nested trace spans counted the same API response twice, that the overstatement is not uniform across rows and that costs have not yet been re-derived. The airline page marks the Generalist rows as a lower bound that is not directly comparable with the Tool-Calling rows, and its trace notice tells users to strip the task's reference actions before treating the recorded actions as the agent's, adding that in some records these actions sit in the interior with agent actions after them. The archives themselves are unchanged, and the maintainers are following the \taub{} issue for the root cause. As of 3 November 2026 the \taub{} issue and the \tautwo{} issue have no maintainer reply, and the community fix for the solo-mode check is open and unmerged. A five-week turnaround with full confirmation is fast for any project, and the outcome is still notes beside numbers that remain as they were.

\section{The common pattern}
\label{sec:pattern}

All four errors were silent in the score they produced. None raised an exception, none produced a value outside the range a reader expects, and none could be detected from the score alone. Each was found only by recomputing the score from lower-level records and comparing.

Each error also arose from a locally sensible decision. Summing usage over all spans is the natural way to catch every call. Computing the reward only when an episode ends is the natural definition of an episode-level reward. Replaying gold through the real environment is the most faithful way to build a reference state. Making replay lenient keeps old files scorable after tools change. The failures came from interactions between components that were each reasonable, written by different people at different times: a tracing library and a client library, a benchmark's termination rule and a scaffold's closing turn, a grader and an archiver, an evaluator and an older file format. Code review of any one component would likely have passed it.

This is why we expect more such errors in agent leaderboards that nobody has re-derived. The expectation is an argument from mechanism. Every interactive benchmark gates grading on termination, and every new scaffold brings its own way of ending. Every leaderboard that logs through a wrapper library over a provider client can nest spans. Every stateful grader that replays gold through the agent's environment can write into the agent's record. Every benchmark that revises its grader while keeping old results faces the version problem. None of these conditions is specific to HAL or \tautwo{}. The search that found these four was also small: two leaderboards, read from their public artifacts over a few days, with no agent runs and no model calls. Our earlier work found the same separation of instrument from model elsewhere, in a multi-turn function-calling benchmark whose scorer never looks at the turn on which the agent should ask \citep{liu2026asking}, and in coding agents, where switching harnesses flips as many tasks as rerunning the same configuration, so a single pass rate is the wrong instrument for telling harnesses apart \citep{liu2026harness}. The conditions that produced them are common, finding them is cheap, and leaderboards are rarely re-derived, which is the situation in which undiscovered cases are expected to outnumber discovered ones.

\section{Checks that expose the instrument}
\label{sec:todo}

Each of the four errors would have been visible from one extra number that the pipeline already had. The checks below cost a few lines of code each and require no change to any task or reward rule.

Report the graded fraction. Every leaderboard cell should state how many of its episodes reached the grader, with the rest counted separately as ungraded. A Generalist cell reading ``accuracy 0.30, graded 51\%'' would have stopped a reader before any conclusion was drawn. Rollout Cards proposes declaring dropped and failed runs \citep{masters2026rollout}, and the graded fraction is the same idea applied inside a run.

Aggregate cost from root spans and print the multiplier. Cost and tokens should be summed over root spans only, or deduplicated by response identifier, and the pipeline should report the ratio of all-span to root-span usage per run. A ratio of exactly two is a bug signature that any reader can recognise, and printing it costs one division.

Tag the provenance of every recorded action. Each logged action should carry who produced it: the agent, the user simulator or the grader. The grader's reference replay should run in a separate environment or write to a separate list. With provenance in the record, a downstream study can filter, and an archive that lacks it cannot be cleaned without re-deriving the grader. Without the tag a reader has to infer provenance from where an action sits, and that inference fails exactly where the harness writes after the grader.

Pin evaluator versions and publish re-grade diffs. Every score should name the evaluator build and task-data version that produced it. A re-grade should print the per-task changes it makes and fail when the share of mismatched tool returns in a replay passes a threshold, instead of warning per call and returning a number. A community pull request, still open, makes \tautwo{}'s re-grading command report the mismatch count and adds the threshold as an option \citep{tau2bench2026pr573}. When a leaderboard is re-graded, the before-and-after table belongs next to it, as \tautwo{} already did for banking. Its 2.0.0 release goes further and keeps 1.x and 2.0 results on separate leaderboard tabs under a warning that they are not comparable \citep{tau2repo}. A recent survey recommends version control for every benchmark component \citep{chen2026emerging}, and this check is that recommendation applied to the grader, with the re-grade itself treated as a measurement that needs its own diff.

Attach errata to archives where they travel. A frozen archive cannot be corrected, but it can carry a machine-readable errata file inside the dataset, listing the known instrument errors, the affected records and the correction rule. A notice on a web page does not reach someone who downloaded the traces before it appeared or who copied a number into a paper.

\section{Discussion}
\label{sec:alt}

Each error has a small fix, and the HAL maintainers acted quickly, but two things outlast a fix. The published numbers do not change when code is fixed, because the archive is frozen and the leaderboard rows were computed once. More important, the class recurs. Each case came from a reasonable interaction between components, so a fix to one component leaves the conditions in place for the next scaffold or the next release. A recurring class of silent error calls for a check that catches the class, which is what Section~\ref{sec:todo} proposes.

Read as rough guides, leaderboards survive these errors, since none of them changes which model generation is stronger than which. They are not small, though. A factor of two applied to some rows of a cost axis and not others, a 13-point gap between two scaffolds that closes when every episode is graded, and a re-grade that takes a score from 0.99 to 0.17 are differences that a reader of a rough guide would act on. Leaderboards that break results down by scaffold and plot cost against accuracy invite exactly the comparisons these errors corrupt.

The disclosure process worked well in the HAL case, and we credit it without reservation. It still depended on an outside reader decrypting archives and replaying graders, which is not a process a field can rely on. The disclosures also live on web pages, while downloaded trajectories and copied numbers circulate without them. A system in which instrument errors surface only when a volunteer re-derives the pipeline will find a small fraction of them.

Releasing code and trajectories made this audit possible, and we would not have found anything without them. Re-running released code reproduces released errors, however, and a trajectory archive that mixes grader and agent actions is reproducible and wrong at once. Openness is necessary for auditing the instrument and does not perform the audit.

\section{Related work}

Kapoor et al.\ argued that agent evaluation should account for cost, guard against overfitting through proper holdouts and standardize its practice for reproducibility \citep{kapoor2024agents}, and HAL implemented a shared harness and published logs at scale \citep{kapoor2025hal}. Our cases are drawn from that infrastructure.

The closest statement of our claim comes from the same group. Their October 2026 survey of agent evaluation observes that the harnesses, autograders and sandboxed environments an agent evaluation requires ``are themselves new surfaces for scrutiny'', warns that a widely adopted framework concentrates the influence of its default metrics, and recommends version control for each benchmark component so that results stay comparable over time \citep{chen2026emerging}. The claim in Section~\ref{sec:exposed} sharpens that view and does not originate it. What this paper adds is evidence that the risk is realized in maintained public leaderboards and is large enough to reorder them, in the form of four silent errors with measured effects on scores and rankings, all located in the run-time pipeline and none in benchmark design. Of our checks, pinned evaluator versions follow the survey's recommendation on versioning, while the graded fraction, the root-span multiplier, action provenance and errata inside archives answer failure modes that its practices do not cover. The survey cites HAL's standardized cost reporting as an example of shared infrastructure, and that cost column is the one our first case finds doubled for some rows, which is what we mean by an instrument that is standardized and still wrong.

Zhu et al.\ introduced the Agentic Benchmark Checklist, which audits task setup and reward design and reports, among other issues, that TAU-bench counts empty responses as successful \citep{zhu2025abc}. UTBoost showed that insufficient test cases in SWE-bench let wrong patches pass and change leaderboard rankings once tests are added \citep{yu2025utboost}. Both examine whether the benchmark's grading rule is right. We take the rule as given and examine whether the published run-time pipeline applied it, logged it, priced it and stored it correctly after release. The two audits are complementary, and a benchmark can pass the checklist while its leaderboard carries every error in Section~\ref{sec:cases}.

Kirgis et al.\ argue that log analysis is necessary for credible agent evaluation and illustrate it on TAU-bench Airline \citep{kirgis2026log}. Our third case adds a precondition: logs must say who wrote each entry, since a grader that writes into the agent's record turns log analysis into partial analysis of the grader. Rollout Cards treats rollout records as the unit of reproducibility, finds that none of 50 audited repositories reports failed or skipped runs, and shows that changing only the reporting rule on fixed rollouts can reverse rankings \citep{masters2026rollout}. Their concern is the rule applied to a fixed record. Ours is that the record and the instrument that produced it can themselves be wrong, and that the instrument changes across versions. \taub{} and \tautwo{} are the benchmarks behind three of our cases \citep{yao2024taubench,barres2025tau2}, and \tautwo{}'s practice of publishing before-and-after leaderboard tables for its banking re-grade is the model for the re-grade diffs we propose. Our own earlier papers report further separations of instrument from model, in multi-turn function calling \citep{liu2026asking} and in coding-agent harnesses \citep{liu2026harness}.

\section{Limitations}

Our evidence comes from two leaderboards and the benchmarks they host, and the expectation that similar errors are common elsewhere rests on the mechanisms in Section~\ref{sec:pattern} rather than on a survey. The HAL re-grades use the \taub{} commit HAL's setup pins with the user simulator stubbed, so they evaluate the database and output criteria only.

\section{Conclusion}

Agent evaluation has learned to ask which model and which scaffold produced a score. The cases here show that it must also ask which instrument produced it, because the instrument is built from fast-moving parts whose interactions fail without a sound. A grader that did not run, a cost that was counted twice, an action the agent never took and a re-grade that rebuilt the wrong world all looked like ordinary results. The fixes are small, and the checks that would have exposed each error are smaller. What the field needs is the habit of publishing them next to every score, so that the next silent error is caught by the leaderboard itself instead of by an outside reader re-deriving its archives.

\section*{Acknowledgements}
We thank the HAL maintainers for reproducing and confirming the three HAL findings on their own archive and for adding disclosures to the leaderboard, and the \tautwo{} contributor who proposed fixes for the solo-mode check and the lenient replay.

\bibliographystyle{plainnat}
\bibliography{refs}

\appendix
\section{How the numbers were obtained}
\label{app:method}

HAL publishes 380 encrypted trace archives on Hugging Face, and the decryption script on its website uses a fixed passphrase. We decrypted the 138 archives for TAU-bench Airline, ScienceAgentBench and CORE-Bench Hard, plus the SciCode and SWE-bench Verified Mini archives used for the cost analysis.

The cost case compares, for each run, the archive's reported token usage with the sum over all logged calls and with the sum over root calls only. In 133 of the 134 runs with logs the reported usage equals the all-calls sum, which identifies the aggregation rule directly. Over the 133 runs whose calls carry usage data the reported cost exceeds the root-span cost by 36.5\%, and one CORE-Bench run whose calls carry no usage block is excluded because neither sum can be formed.

For the TAU-bench cases we installed the \taub{} commit that HAL's benchmark setup pins and replayed recorded actions through the environment with the language-model client and the user simulator stubbed, so that grading evaluates the database comparison and the output check without any model call. To separate the grader's actions from the agent's, we parsed the agent's own tool calls from the model outputs logged in each archive and aligned them, in order, against the recorded action list by tool name and arguments. Recorded actions with no matching agent call are the grader's. An episode counts as graded when its record holds such an injected copy of the gold actions anywhere, or when its recorded reward is positive. Our first audit, reported on the HAL discussion page in September, used a simpler rule, counting an episode as graded only when the gold actions formed the end of the record, and that rule missed the 73 Generalist episodes whose only injected copy sits in the middle of the record. The figures here supersede that report, and a correction is posted in the same discussion \citep{hf2026discussion9}. HAL installs its agents from a fork whose airline gold differs from the pinned commit on three of the 50 tasks, so every re-grade was also run against the gold embedded in each record, which leaves the pooled figures unchanged.

Two checks support the classification. Re-running the grader at each recorded ending event, with the gold embedded in the record, reproduces the recorded reward for all 1,951 graded records. Replaying only the agent's own actions on graded episodes reproduces the recorded reward for 97\% of Tool-Calling, 97\% of Few-Shot and 84\% of Generalist records, and almost every Generalist disagreement is an episode the grader passed that the offline replay fails, so the offline re-grade of ungraded episodes, if anything, understates the Generalist. The pooled figures in the main text pool all runs of each scaffold over records whose gold contains at least one non-terminal action. Table~\ref{tab:permodel} gives the per-model values behind Figure~\ref{fig:regrade}. Including the withdrawn Few-Shot traces, the Generalist is last of the three scaffolds for nine of the ten models with a second scaffold as published, and last for six once every episode is graded, while it is first for claude-3.7-sonnet, ties Few-Shot for first on gemini-2.0-flash, ties it for last on deepseek-r1 and is second on deepseek-v3.

\begin{table}[h]
\centering
\small
\begin{tabular}{lccccc}
\toprule
& Generalist & Tool- & Generalist, & Generalist, & Few-Shot \\
Model & records & Calling & published & all graded & (withdrawn) \\
\midrule
claude-3.7-sonnet & 81 & 0.49 & 0.52 & 0.60 & 0.46 \\
claude-opus-4 & 82 & n/a & 0.41 & 0.50 & 0.57 \\
claude-opus-4.1 & 82 & 0.51 & 0.37 & 0.50 & 0.55 \\
deepseek-r1 & 41 & 0.31 & 0.10 & 0.24 & 0.24 \\
deepseek-v3 & 41 & 0.38 & 0.17 & 0.37 & 0.22 \\
gemini-2.0-flash & 41 & 0.22 & 0.20 & 0.32 & 0.32 \\
gpt-4.1 & 41 & 0.29 & 0.20 & 0.27 & 0.51 \\
gpt-5 & 82 & 0.44 & 0.34 & 0.43 & 0.50 \\
o3 & 40 & 0.49 & 0.20 & 0.38 & 0.46 \\
o4-mini & 82 & 0.41 & 0.18 & 0.38 & 0.49 \\
\bottomrule
\end{tabular}
\caption{Per-model TAU-bench Airline accuracy on HAL, records whose gold contains at least one non-terminal action, pooled over each model's runs. Tool-Calling and Few-Shot are shown after re-grading, which leaves Few-Shot unchanged and moves Tool-Calling only for claude-3.7-sonnet (published 0.48) and gpt-4.1 (published 0.27). HAL withdrew Few-Shot for test-set leakage, and its archived traces are shown for reference. claude-opus-4 has no Tool-Calling run.}
\label{tab:permodel}
\end{table}

For \tautwo{} we installed four builds side by side (0.2.0, 1.0.0, the commit before 1.0.1, and 1.0.1) and re-graded public trajectory files under each, varying evaluator code and task data independently. The fidelity check is exact: our 1.0.1 re-grade of claude-opus-4-5 on banking gives a pass rate of 0.2474 against the official 24.74\%. Banking was re-graded on five leaderboard files that parse under every build, while the 17-submission figures come from the repository's own before-and-after submission files. Code, commit identifiers and per-run outputs are in the release \citep{liu2026release}.

\end{document}